\documentclass{article}

\usepackage{microtype}
\usepackage{graphicx}
\usepackage{subfigure}
\usepackage{booktabs} 
\usepackage{outlines}
\usepackage{xcolor}
\usepackage{amssymb}
\usepackage{pifont}
\usepackage{amsthm}
\usepackage{soul}
\usepackage{blindtext}
\usepackage{titlesec}
\usepackage{listings}
\usepackage{enumitem}
\usepackage{multirow}

\usepackage[most]{tcolorbox}

\newtcolorbox{codebox}{
  colback=gray!5,    
  colframe=black!60, 
  boxrule=0.5pt,     
  arc=4pt,           
  boxsep=0pt,        
  left=4.0pt,
  right=4.0pt,
  top=0.0pt,
  bottom=0.0pt,
}

\theoremstyle{definition}
\newtheorem{definition}{Definition}[section]

\newif\ifShowComments

\definecolor{OliveGreen}{cmyk}{0.64,0,0.95,0.40}

\usepackage{hyperref}

\usepackage[accepted]{icml2024}

\usepackage{amsmath}
\usepackage{amssymb}
\usepackage{mathtools}
\usepackage{amsthm}
\usepackage[hang,flushmargin]{footmisc}
\usepackage{listings}
\usepackage{xcolor}

\definecolor{codegreen}{rgb}{0,0.6,0}
\definecolor{codegray}{rgb}{0.5,0.5,0.5}
\definecolor{codepurple}{rgb}{0.58,0,0.82}
\definecolor{backcolour}{rgb}{0.95,0.95,0.92}

\lstdefinestyle{mystyle}{
    backgroundcolor=\color{backcolour},
    commentstyle=\color{codegreen},
    keywordstyle=\color{magenta},
    numberstyle=\tiny\color{codegray},
    stringstyle=\color{codepurple},
    basicstyle=\ttfamily\footnotesize,
    breakatwhitespace=false,
    breaklines=true,
    captionpos=b,
    keepspaces=true,
    numbers=left,
    numbersep=5pt,
    showspaces=false,
    showstringspaces=false,
    showtabs=false,
    tabsize=2,
    basicstyle=\ttfamily\scriptsize,
    language=Python,
    morekeywords=[2]{self.cur_targets, self.prev_targets, torque, self.actions},
    keywordstyle=[2]\color{codepurple},
}

\usepackage[capitalize,noabbrev]{cleveref}

\theoremstyle{plain}

\usepackage[textsize=tiny]{todonotes}

\icmltitlerunning{
Position: Automatic Environment Shaping is the Next Frontier in RL
}

\begin{document}

\twocolumn[
\icmltitle{
Position: Automatic Environment Shaping is the Next Frontier in RL
}



\icmlsetsymbol{equal}{*}

\begin{icmlauthorlist}
\icmlauthor{Younghyo Park}{equal,mit}
\icmlauthor{Gabriel B. Margolis}{equal,mit}
\icmlauthor{Pulkit Agrawal}{mit}
\end{icmlauthorlist}

\icmlaffiliation{mit}{Improbable AI Lab, Massachusetts Institute of Technology, Cambridge, MA, USA}

\icmlcorrespondingauthor{Younghyo Park}{younghyo@mit.edu}
\icmlcorrespondingauthor{Gabriel B. Margolis}{gmargo@mit.edu}

\icmlkeywords{Machine Learning, ICML}

\vskip 0.3in
]
\printAffiliationsAndNotice{\icmlEqualContribution} 

\begin{abstract}

Many roboticists dream of presenting a robot with a task in the evening and returning the next morning to find the robot capable of solving the task. What is preventing us from achieving this? Sim-to-real reinforcement learning (RL) has achieved impressive performance on challenging robotics tasks, but requires substantial human effort to set up the task in a way that is amenable to RL. It's our position that algorithmic improvements in policy optimization and other ideas should be guided towards resolving the primary bottleneck of shaping the training environment, i.e., designing observations, actions, rewards and simulation dynamics. Most practitioners don't tune the RL algorithm, but other environment parameters to obtain a desirable controller. We posit that scaling RL to diverse robotic tasks will only be achieved if the community focuses on automating environment shaping procedures. 
\end{abstract}

\section{Introduction}
\label{sec:intro}

\begin{figure*}[t!]
\centering
\includegraphics[width=0.95\textwidth]{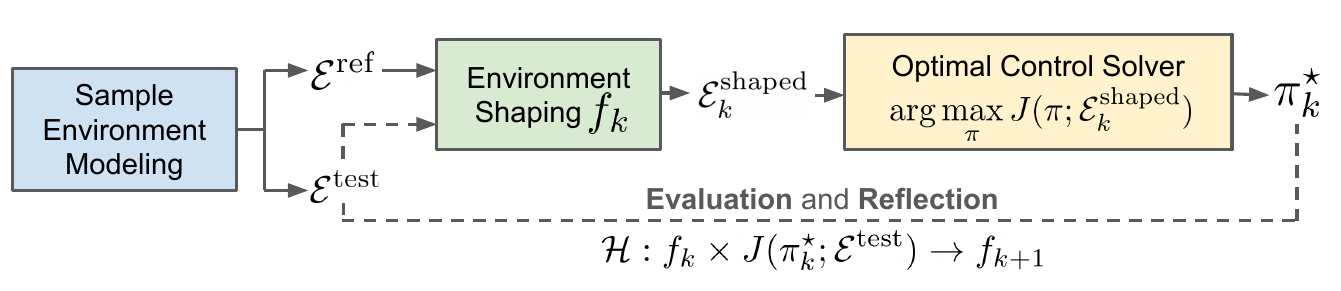}
\caption{Flowchart of a typical behavior generation pipeline using reinforcement learning with simulation, illustrating four distinct subtasks of sample environment modeling, environment shaping, RL training, and outer feedback loop with behavior evaluation and reflection. We highlight the manual, task-driven \textbf{environment shaping} as a key, yet often overlooked, bottleneck in generalizing the success of RL. We thus advocate for automating the environment shaping process to broaden RL's applicability.}
\label{fig:main-figure}
\end{figure*}

The advent of foundation models for speech, vision, and language processing has revolutionized Artificial Intelligence. It is widely believed that robotics is the next frontier, and the race to develop a foundation model for robotics is ongoing.  The critical challenge in this pursuit is the limited availability of robotic data: trajectories of observations and robot actions. This starkly contrasts with computer vision and natural language processing, where large amounts of readily available internet data can be used. One way to gather robotic data is to deploy robots for many tasks in diverse environments. However, such deployment is only feasible if robots create value, i.e., successfully solve the task often enough. The result is a chicken-and-the-egg situation -- robots need to be useful to be deployed, but for them to be useful requires collecting enough data to train a  controller that creates value. The real question, therefore, is bootstrapping data collection. 

A natural way to collect data is by teleoperating a robot to perform diverse tasks. However, this paradigm is challenging to scale as the human effort grows linearly with the amount of data that needs to be gathered. To ease data collection, recent works have made teleoperation more capable and easier~\cite{ALOHA, MobileALOHA}, but it doesn't change the linear scaling.  At some point in the future, it is plausible that we will have enough data to train a large model that will reduce the number of demonstrations required to learn a new task and make the human effort sub-linear. However, we are far from that point. The primary reason is that policies obtained via supervised learning on a small dataset of demonstrations (i.e., learning from demonstration) have limited robustness and generalization and, therefore, cannot be used to collect data autonomously in more diverse settings.   

In theory, given a reward function or by inferring a reward function from the demonstrations, autonomous data collection is possible using reinforcement learning (RL). It has been hard to realize this promise of RL because training in the real world often requires babysitting the robot to ensure safe operation, ensuring the reward function is not hacked, performing resets, and the data inefficiency of RL algorithms means they run for a long time before useful behaviors are discovered. Real-world RL training is an active area of research~\cite{luo2024serl}, and recent work has shown the plausibility of learning locomotion behaviors from real-world RL training~\cite{wu2023daydreamer, lei2023uni}. However, real-world RL is yet to achieve state-of-the-art robotic controllers, which means the data it generates is sub-optimal. 

A related line of work bypasses the difficulty of training in the real world by training with RL in simulation and then successfully deploying policies in reality (i.e., sim-to-real RL). Such training has achieved state-of-the-art behaviors across many robot morphologies and complex tasks involving legged locomotion \cite{miki2022learning, ji2023dribble, hoeller2023anymal, zhuang2023robot, cheng2023extreme, jenelten2024dtc}, dexterous manipulation \cite{andrychowicz2020learning, chen2023visual, handa2023dextreme, yin2023rotating}, drone racing~\cite{song2023reaching, kaufmann2023champion} and others. A single general-purpose RL algorithm, PPO~\cite{schulman2017proximal}, has powered these successes. Consequently, one might believe that a recipe for scaling to diverse tasks and collecting a sizeable high-quality dataset of trajectories exists. However, the reality is that immense manual task-specific modeling and engineering are required -- something we call \textit{environment shaping} (a generalization of the term reward shaping to include other environment choices to ease optimization) --  to make things work, which is the primary bottleneck in our opinion. Although the environment shaping bottleneck is highly relevant to robotics, it exists in any domain where reinforcement learning is applied to real-world problems, including transportation systems~\cite{wei2018intellilight}, autonomous driving~\cite{wu2017emergent}, finance~\cite{finrl_meta_2021}, and power management~\cite{vazquez2019citylearn}.

We expand the previous usage of the term environment shaping~\citep{co2020ecological} to include all choices such as designing reward function, curriculum, observation/action spaces, initial state distribution, reset functions performed manually to make training possible. These issues have been individually studied for a long time~\cite{selfridge1985training, ng1999policy, singh2009rewards, esser2023guided}, but a holistic and critical analysis of the human effort required to make these choices even with the latest algorithmic advances has not been made.

Using examples from recent applications of RL to robotics,  \textbf{this position paper argues that the primary bottleneck for scaling up reinforcement learning is its need for manual environment shaping.} Specifically, this is a call to action for the RL research community:

\begin{itemize}[leftmargin=*, topsep=0pt, partopsep=0pt, itemsep=-2pt]
    \item Distinguish \textit{modeling} from \textit{shaping} design choices in RL environments. Many works describe the final shaped environment but not a repeatable procedure that can transfer the same shaping to a new robot or task.
\end{itemize}
\begin{itemize}[leftmargin=*, topsep=0pt, partopsep=0pt, itemsep=-2pt]
    \item Prioritize research into \textit{automatic environment shaping} as a pathway to generalize domain-specific successes in RL.
\end{itemize}
\begin{itemize}[leftmargin=*, topsep=0pt, partopsep=0pt, itemsep=-2pt]
    \item Prioritize \textit{benchmarks} that measure the total expense of applying reinforcement learning to real-world robotics tasks, by including environments with `unshaped' versions and corresponding human-shaped baselines. The point being that existing benchmarks have already performed environment shaping (i.e., hide the true problem) and, therefore do not serve as good candidates for further research into better algorithms. \footnote{~\footnotesize \url{https://auto-env-shaping.github.io/}}
\end{itemize}

\section{Robotic Behavior Generation with RL}

To support a precise definition of environment shaping, we first describe a typical workflow for generating robotic behaviors using reinforcement learning (RL) in simulation (Figure \ref{fig:main-figure}). 

We decompose behavior generation into four subtasks; sample environment generation (Sec \ref{sec:sample-env-gen}), environment shaping (Sec \ref{sec:env-shaping}), RL training (Sec \ref{sec:rl-training}), and the outer feedback loop with behavior evaluation and reflection (Sec \ref{sec:behavior-eval}). In delineating these substages, we pinpoint typical human efforts involved in the process.

\subsection{Modeling Sample Environments}\label{sec:sample-env-gen}

Consider $\hat{p}(e)$ as the oracle distribution of the target environment we want to deploy our robots in. Our goal is to generate behavior that is performant (with respect to objective $J$) and robust under $\hat{p}(e)$, 
$$
\max_\pi \mathbb{E}_{\hat{e} \sim \hat{p}(e)} J(\pi; \hat{e})
$$
The target environment can either be a specific real-world environment that already exists (e.g. kitchen at a specific location) or a generic concept (e.g. typical household kitchen). 

\begin{figure}[t]
\centering
\includegraphics[width=0.9\columnwidth]{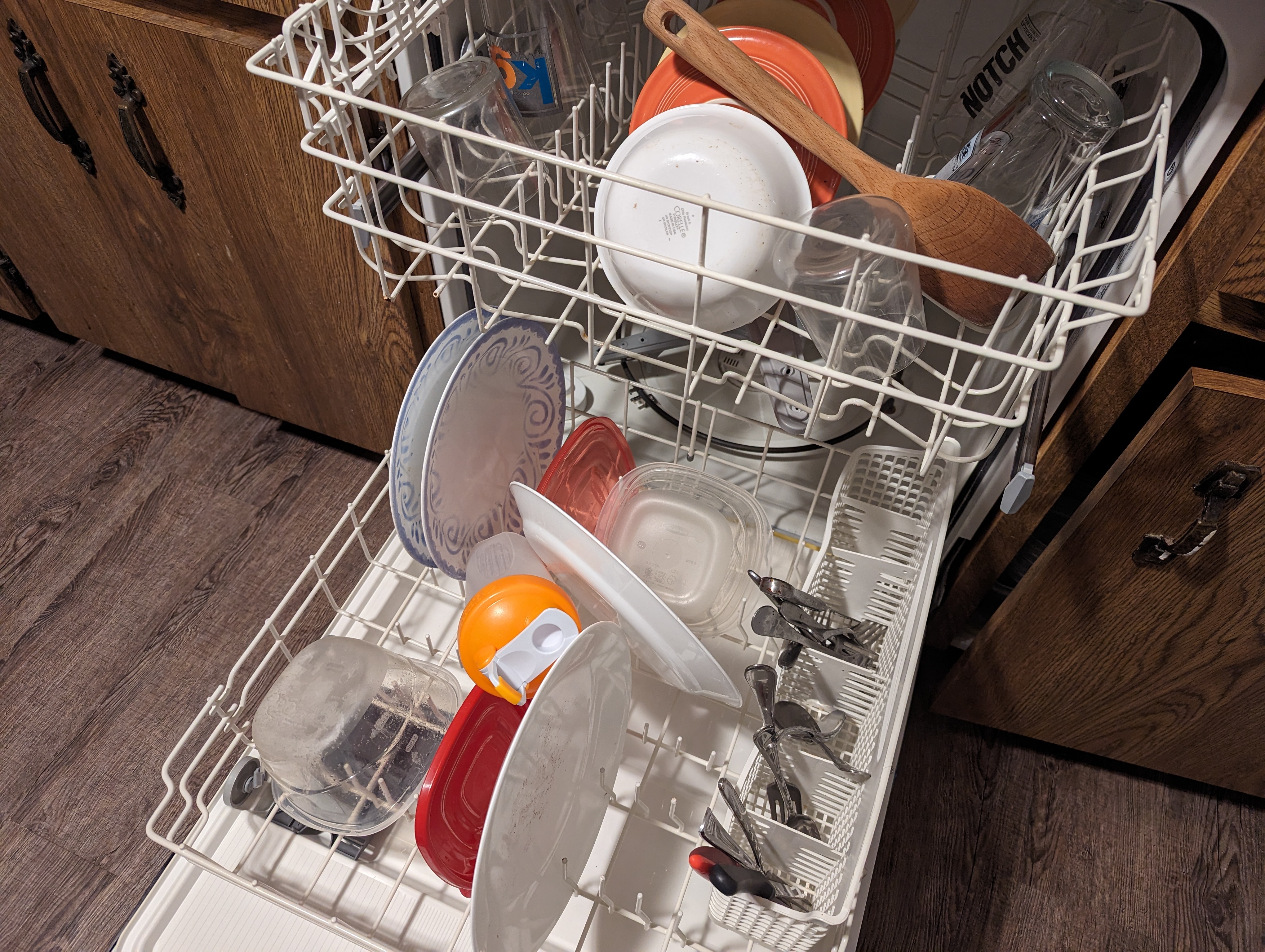}
\caption{Example of environment complexity: an overloaded and disorganized real-world dishwasher.}
\label{dishwasher}
\end{figure}

Unfortunately, it's extremely difficult to model this oracle distribution either way; it requires comprehensive knowledge of all possible environmental variables and conditions, which is often infeasible due to the complexity, variability, and limited observability in the real world. Innate vagueness of generic concepts is often an issue as well. Just imagine modeling a true oracle distribution of a chaotic real-world dishwasher in simulation! (Figure \ref{dishwasher})

In contrast, modeling a \textit{single} sample environment, $\hat{e}$, a specific instance drawn from oracle environment distribution, 
$$
\hat{e} \sim \hat{p}(e),
$$
and importing that to a simulation is much more feasible. This is why robotics practitioners typically start the behavior generation process by first designing a single sample environment: (a) modeling and importing robots and necessary assets in simulation, and (b) manually placing them in their default poses. We often generate a \textit{set} of those sample environments to kick things off. This is what practitioners do for the first blue box in Figure \ref{fig:main-figure}.

Such a set of sample environments actually serves a purpose: it is a useful representative testbed environment that can be used to estimate the behavior performance under the true oracle distribution $\hat{p}(e)$. Combined with any form of task specification $r$ \cite{task-specification-problem}, we define a simulated testbed $\mathcal{E}^{\text{test}}$ where the trained behaviors can be evaluated.

\begin{definition}[Test Environment] Let $\{ \hat{e}_1, \cdots, \hat{e}_n\}$ be a set of $n$ \textit{sample environments} each independently drawn from oracle environment distribution $\hat{p}(e)$. The simulated counterparts are denoted as  $\hat{e}_{\text{sim}, i}$. Given a task specification $r$, a test environment ${\mathcal{E}}^{\text{test}}$ is defined as a set of tuples: 
$$
\mathcal{E}^{\text{test}} = \{\left < \hat{e}_{\text{sim}, i}, r \right > \}_{i=1, \cdots, n}, 
$$
where the generated behavior $\pi$ will be evaluated in.
\end{definition}

Meanwhile, to prevent the behavior from overfitting to a few sample environments within $\mathcal{E}^{\text{test}}$, one can maintain two sets of sample environments; the test environment can be strictly held out from the rest of the behavior generation pipeline, letting it serve the sole purpose of behavior evaluation. A distinct set of sample environments then can be used to effectively guide the remaining design choices. We call that a \textit{reference} environment, $\mathcal{E}^{\text{ref}}$, as illustrated in Figure \ref{fig:main-figure}.

\begin{definition}[Reference Environment] A Reference Environment $\mathcal{E}^{\text{ref}}$ is a distinct set of sample environments that provides useful context to the shaping algorithm. Trained behaviors will not be evaluated here to avoid overfitting. 
\end{definition}

For instance, when designing a robotic behavior for unloading a dishwasher, sample environments would include multiple instances of dishwashers loaded with varying configurations of dishes and utensils. Sampled configurations then could be split into a set of reference environments for guiding the shaping and a set of held-out test environments for evaluation. This diversity in configurations provides the shaping algorithm with a broad context, incentivizing it to infer the underlying distribution of object placements. The goal is for the reinforcement learning (RL) trainer to sample from this inferred distribution during training, ensuring that the generated behaviors are robust and adaptable to various real-world scenarios without overfitting to a specific set of environments.

\subsection{Shaping Reference Environments}\label{sec:env-shaping}

A straightforward subsequent step of behavior generation might be to directly use the reference environments $\mathcal{E}^{\text{ref}}$ as an RL environment, with the expectation that algorithms like Proximal Policy Optimization (PPO) \cite{schulman2017proximal} will find performant and generalizable behaviors. However, this approach often falls short due to the inherent sparsity of these environments.

Reference environments often present a challenging optimization landscape for RL algorithms due to their \textit{sparse} nature. 
For example, it might be rare to obtain nonzero rewards or the observation space might be dominated by spurious features that encourage poor local minima. 
To mitigate these challenges, human engineers typically go through a process of \textit{shaping} the elements of $\mathcal{E}^{\text{ref}}$. This modification, aimed at enhancing the \textit{learnability} of the environment, includes introducing denser learning signals and additional modifications encouraging effective exploration. 
The resulting \textit{shaped environment} \cite{co2020ecological} then used as a training ground for the subsequent optimal control solver.

\begin{definition}[Shaped Environment]
A \textit{Shaped Environment} ${\mathcal{E}}^{\text{shaped}}$ is a modification of a reference environment, i.e., ${\mathcal{E}^{\text{shaped}}} = f(\mathcal{E}^{\text{ref}})$. The transformation $f$ incorporates design choices specifically optimized for learning performance, smoothing the optimization landscape enabling the solver to find better solutions with maximal performance in $\mathcal{E}^{\text{test}}$. The transformation function 
\begin{equation}
f: \mathcal{E}^{\text{ref}} \rightarrow \mathcal{E}^{\text{shaped}}
\end{equation}
is defined as \textit{shaping function}.
\end{definition}

The \textit{shaping} process for a robotics environment usually involves manually explored design choices, including shaping the reward \cite{ng1999policy, singh2009rewards}, modifying the action space \cite{peng2017learning, aljalbout2023role}, designing curricula on the environment dynamics, initial state distributions, and goal distributions \cite{selfridge1985training, matiisen2019teacher, portelas2020automatic, lee2020learning}, crafting the state space \cite{yu2023identifying}, and shaping the right failure conditions for early termination \cite{co2020ecological}. We describe details of common shaping operations in Sec \ref{sec:isaacgymenvs-analysis}.  

\subsection{RL Training} \label{sec:rl-training}

Once the shaped environment is obtained, the next step of behavior generation is to use RL algorithms, i.e., PPO \cite{schulman2017proximal}, to find a behavior $\pi$ that best performs on the shaped environment $\mathcal{E}^{\text{shaped}}$. Formally, the algorithm aims to find an optimal behavior $\pi$ for the following optimization problem: 
\begin{equation}
\begin{aligned}
& \max_{\pi} \mathbb{E}_{\tau \sim \pi} \left[ \sum_{t=0}^{T} \gamma^t r_t (s_t, a_t)) \right] \\
& ~~ \text{s.t. } ~~ s_{t+1} \sim p(s_t, a_t; \mathcal{E}^{\text{shaped}}).
\end{aligned}
\end{equation}
While the RL training process also requires a range of design decisions, such as algorithmic choices and hyperparameter adjustments, these areas are relatively well-researched and documented \cite{parker2022automated, kiran2022hyperparameter}. 

However, we note that in a pratical context of robotic behavior generation, tuning the RL setting (e.g. neural architecture search for policy or hyperparameter tuning) is often underprioritized compared to the effort put into environment shaping. In IsaacGymEnvs \cite{makoviychuk2021isaac}, for instance, simple Multilayer Perceptron (MLP) networks are employed, and off-the-shelf RL algorithm implementations are utilized with their default configurations. 
This shows that algorithms like PPO and their default settings are capable enough when paired with \textit{ideally shaped} environment.

\subsection{Optimizing Environment Shaping via Iterative Behavior Evaluation and Reflection}\label{sec:behavior-eval}

Once an optimal behavior $\pi^\star$ is obtained via RL training, the behavior is \textit{evaluated} on the test environment $\mathcal{E}^{\text{test}}$ and \textit{reflected} by human engineers. Denoting the reflection process as $\mathcal{H}$ of analyzing the generated behavior $\pi^\star$ in test environment $\mathcal{E}^{\text{test}}$ and coming up with a better environment shaping $f$, 
\begin{equation}
\mathcal{H}:  f_k \times J(\pi_k^\star; \mathcal{E}^{\text{test}}) \rightarrow f_{k+1},
\end{equation}

robotic behavior generation process can be formally defined as an \textbf{iterative optimization process} over the \textbf{environment shaping function} $f$, 
\begin{equation}
\begin{aligned}
& f_{k+1} = \mathcal{H} \left ( f_k, J(\pi^\star_k; \mathcal{E}^{\text{test}})\right ) \\
& \text{where} ~~ \pi^\star_k = \underset{\pi}{\text{argmax}} \ J(\pi; \mathcal{E}^{\text{shaped}}_k), \\
& ~~~~~~~~~~~~ \mathcal{E}^{\text{shaped}}_k = f_{k}(\mathcal{E}^{\text{ref}}), ~~ f_0 = \mathbf{I}_{\text{identity}} 
\end{aligned}
\end{equation}
which aims to find an optimal shaping function $f \in \mathcal{F}$ for the following bi-level optimization problem: 
\begin{equation}
\begin{aligned} 
& {\color{black}\max_{f \in \mathcal{F}}} ~ J(\pi^\star; \mathcal{E}^{\text{test}}) \\ & ~\text{s.t.} ~~~ \pi^\star \in \arg\max_\pi J({\pi};{\mathcal{E}^{\text{shaped}}}), ~~{\mathcal{E}^{\text{shaped}}} = {\color{black} f}(\mathcal{E}^{\text{ref}}).
\end{aligned}
\end{equation}

After shaping is applied, the new environment may not reflect the original task; therefore, note that the outer level of the bilevel optimization maximizes $J(\pi^*; \mathcal{E}^\mathrm{test})$ which is the return evaluated in the original test environments without any shaping. If the inner level optimizes a shaped environment well, but with poor correspondence to the original task, it will be dispreferred by the outer loop.

\section{The Current State of Environment Shaping}

Having established the role of environment shaping operations in successful behavior generation, we now probe deeper into the unique challenges that the problem of environment shaping and its optimization procedure presents. 
Specifically, we make the following arguments with supporting experiments and analysis:
\begin{itemize}[leftmargin=*, topsep=0pt, partopsep=0pt, itemsep=0pt]
\item Popular RL benchmark environments are \textit{artificially} made easy for RL with task-specific environment shaping. We should benchmark our algorithms in unshaped environments if we want them to solve new problems without task-specific environment shaping step. (Section \ref{sec:isaacgymenvs-analysis})
\item Shaping multiple attributes of the environment at once (reward, observation space, action space, etc.) is a tricky, non-convex optimization problem (Section \ref{sec:localoptima}).
\item Reward shaping is not the only problem. Existing automation efforts focus too narrowly on rewards (Section \ref{sec:eval-eureka}).
\end{itemize}

\subsection{RL Benchmarks for Robotics are Artificially Easy} 
\label{sec:isaacgymenvs-analysis}

Benchmark environments for robot reinforcement learning include task-specific environment shaping to make it feasible to help baseline RL algorithms perform reasonably well. However, these modified environments might not fully assess how RL algorithms progress towards solving various control problems independently, treated as a black-box, without needing task-specific adjustments.

In this section, we outline the common task-specific design choices associated with different aspects of environment shaping, while formally defining what an \textit{unshaped} counterpart --- one with minimal or no human engineering required --- would look like. We consider a case study of the IsaacGymEnvs task suite~\cite{makoviychuk2021isaac}.

\begin{table}[t]
\centering
\footnotesize
\begin{tabular}{@{}lrl@{}}
\toprule
\textbf{AllegroHand} ~~~~~  & Reward & Change \\ \midrule 
all shaped                       & $38777$  & --     \\
\midrule
sparse reward                       & $0$  & $\downarrow 38777$        \\
unshaped action space               &     $21530$  & $\downarrow 17247$       \\
unshaped observation space       &  $2114$  & $\downarrow 36663$            \\
no early termination           & $0$  & $\downarrow 38777$          \\
single initial state & $0$  & $\downarrow 38777$           \\
single goal state & $141155$  & $\uparrow 102378$          \\ 
\bottomrule
\end{tabular}
\vspace{0.5em}

\centering
\begin{tabular}{@{}lrl@{}}
\toprule
\textbf{Humanoid} ~~~~~~~~~~~~~  & Reward & Change \\ \midrule
all shaped                       & $7554$   & --    \\
\midrule
sparse reward                       & $5237$  & $\downarrow 2317$         \\
unshaped action space               &  $67$ & $\downarrow 7487$     \\
unshaped observation space ~~       &  $0$    & $\downarrow 7554$       \\
no early termination           & $705$  & $\downarrow 6849$      \\
single initial state & $5735$ & $\downarrow 1819$        \\ \bottomrule
\end{tabular}
\vspace{0.5em}

\centering
\begin{tabular}{@{}lrl@{}}
\toprule
\textbf{Anymal} ~~~~~~~  & Reward & Change \\ \midrule 
all shaped                       & $-45$  & --     \\
\midrule
sparse reward                       & $-2789$  &  $\downarrow  2744$      \\
unshaped action space               &     $-2499$   &  $\downarrow  2454$   \\
unshaped observation space ~~       &  $-2656$   &  $\downarrow  2611$        \\
no early termination           & $-43$   &  $\uparrow 2$     \\
single initial state & $-17$   &  $\uparrow 28$  \\
single goal state & $-2516$   &  $\downarrow 2470$        \\ \bottomrule
\end{tabular}
\caption{\textbf{Impact of environment shaping on policy optimization.} Removing task-specific design choices in the reward, action space, state space, early termination, or initialization incurs performance reductions. Top row: environment with original \textit{shaped} design choices. Each subsequent row shows performance after training with a corresponding unshaped design choice. The performance of all policies is evaluated in a fully unshaped environment. }
\label{table:shaping-matters}
\end{table}

\begin{figure}[t] 
    \begin{lstlisting}[language=Python]
 def shaped_action_space(self):
   # scale the targets by the joint limits
   self.cur_targets = scale(self.actions, 
          self.shadow_hand_dof_lower_limits, 
          self.shadow_hand_dof_upper_limits)

   # compute the moving average of targets
   self.cur_targets = self.alpha * self.cur_targets + 
                  (1 - self.alpha) * self.prev_targets
   self.cur_targets = tensor_clamp(self.cur_targets, 
                self.shadow_hand_dof_lower_limits, 
                self.shadow_hand_dof_upper_limits)
   self.prev_targets = self.cur_targets[:]

   # compute the torques according to PD control law
   torque = 3 * (self.cur_targets - self.shadow_hand_dof_pos) - 0.1 * self.shadow_hand_dof_vel

   # apply torques
   self.gym.set_dof_actuation_force_tensor(self.sim, gymtorch.unwrap_tensor(torque))\end{lstlisting}

    \begin{lstlisting}[language=Python]
 def unshaped_action_space(self):
   # directly apply the prediction action as torques
   self.gym.set_dof_actuation_force_tensor(self.sim, gymtorch.unwrap_tensor(self.actions))\end{lstlisting}
  \vspace{-0.2cm}
  \caption{Action space shaping: (Top) Original shaped action space with task-specific features. (Bottom) Unshaped action space consisting of joint torque commands. Some shaped code has been slightly modified from the source to increase brevity and clarity while preserving the original logic.}
  \label{fig:action_shaping_code}
\end{figure}

\begin{figure}[t] 
    \begin{lstlisting}
 def shaped_observation_space(self):
   root_states = rigid_bodies[:, self.root_handle]
   base_quat = root_states[:, 3:7]
   # base linear velocity (from local frame)
   base_lin_vel = quat_rotate_inverse(base_quat, root_states[:, 7:10]) * lin_vel_scale
   # base angular velocity (from local frame)
   base_ang_vel = quat_rotate_inverse(base_quat, root_states[:, 10:13]) * ang_vel_scale
   # transformed base orientation
   projected_grav = quat_rotate(base_quat, gravity_vec)
 
   # scaling/normalizing values
   commands_scaled = scale(commands, 
                  [lin_vel_scale, ang_vel_scale])
   dof_pos_scaled = dof_pos_scale * 
                  (dof_pos - default_dof_pos) 
   dof_vel_scaled = dof_vel_scale * dof_vel

   # concatenate the relevant features
   obs = torch.cat([base_lin_vel, base_ang_vel, 
                   projected_grav, commands_scaled,
                   dof_pos_scaled, dof_vel_scaled, 
                   actions], dim=-1)\end{lstlisting}
    \begin{lstlisting}
 def unshaped_observation_space(self):
    # include entire unprocessed simulator state
    obs = torch.cat([dof_pos, 
                dof_vel, torques, 
                rigid_bodies.reshape(num_envs, -1),
                rigid_body_force.reshape(num_envs, -1),
                commands, actions], dim = -1)\end{lstlisting}
  \vspace{-0.2cm}
  \caption{State space shaping: (Top) Original shaped state space with task-specific features. (Bottom) Unshaped state space contains the entire raw simulator state.}
  \label{fig:state_shaping_code}
\end{figure}

\textbf{Action Space. }~How would an \textit{unshaped} action space look, and how \textit{shaped} is the action space in the case environments? 

For a typical robot in rigid multibody simulation, the unshaped action space would be the motor torques: passing in the policy output $a$ directly to the motor as a torque $\tau$ --- no scaling or transforming the outputs, no gains to be tuned. 

Designing a shaped action space for a robot is thus equivalent to the problem of choosing a low-level controller (and its corresponding parameters) that converts a policy output $\mathbf{a}$ with different physical meanings into an executable motor torque $\tau$ that can drive the actuators. The low-level controller can implement a prior like whether to resist or comply to external forces. Examples include proportional-derivative control, differential inverse kinematics controller, operational space control \cite{khatib1985operational}, or impedance control \cite{hogan1984impedance}. These are projections of the policy outputs onto a strict subset of the space of possible torque commands. The choice of action space can significantly influence the performance of learning algorithms in the environment~\cite{peng2017learning}. 

In IsaacGymEnvs, there are multiple distinct cases of shaped action spaces. \texttt{AllegroHand} uses joint position targets with moving average smoothing as its action space (See Figure \ref{fig:action_shaping_code} for example code). For \texttt{Anymal}, the relative joint position target is used as an action space with proportional-derivative controller. For \texttt{Humanoid}, scaled joint torques are used. Table \ref{table:shaping-matters} empirically shows that removing these \textit{shaping} operations on action space largely impacts the training performance. 

\textbf{Observation Space.}~ What is an \textit{unshaped} observation space, and how did IsaacGymEnvs \textit{shape} the observation? 

Crafting a \textit{shaped} observation space involves strategic selection and transformation of variables from an unshaped observation --- the entire state in simulation that's been exposed to the user. The process, integral to optimizing policy learning performance, includes (a) useful transformation of the unshaped variables (i.e., features) and (b) discarding unnecessary variables. Figure \ref{fig:state_shaping_code} shows the details of the shaped state space in the IsaacGymEnvs \texttt{Anymal} environment. 

Different choices of observation space shaping can substantially impact robot performance. In the setting of quadrupedal locomotion, \citet{yu2023identifying} found that different aspects of the task (robustness, performance) are sensitive to different features included in the observation space. Table \ref{table:shaping-matters} shows our findings agreeing that observation shaping has a significant impact on training performance. For \texttt{Humanoid}, for instance, using an unshaped observation space makes the task completely unlearnable. 

\textbf{Reward. }~The \textit{unshaped} reward represents the true objective function we actually wish to optimize, often defined as {extrinsic} reward \cite{singh2010intrinsically, singh2009rewards} or task-fitness function \cite{niekum2010genetic}.

Shaping the \textit{extrinsic} reward into a more informative \textit{intrinsic} reward to make it more conducive to RL algorithms is a well-studied topic in the community \cite{ng1999policy, singh2010intrinsically, gupta2022unpacking, eschmann2021reward}. The shaping procedure is usally designed to provide \textit{denser} learning signals and to prevent the policy from overly exploiting the innate vagueness of extrinsic reward, i.e., reward hacking. It typically involves a few commonly used strategies. One popular strategy is to reward \textit{progress} towards the goal by adding distance metrics measuring how close the agent is to reaching it. If some attributes of successful behavior are known beforehand, those terms can be added to the shaped reward \cite{chen2020learning, margolis2023walk}. If a trajectory of a successful behavior is known, one can also compute a \textit{similarity} between the generated behavior and the trajectory and try to maximize the similarity \cite{peng2021amp}. Offering bonuses for exploring new states \cite{pathak2017curiosity} is also known to facilitate RL training. Table \ref{table:shaping-matters} shows that \textit{shaping} reward has significant impact on the training performance. 

It's worth noting, however, that defining the extrinsic reward itself poses its own set of challenges, which is well addressed in \cite{task-specification-problem}. The task fundamentally requires translating human intentions and goals into numerical values that an algorithm can optimize, often posing significant challenges. In the scope of this position paper, we presume the task has been defined and consider the challenge of learning a policy to perform it. 

\textbf{Initial/Goal State.} Let's consider a concrete example: how would you \textit{shape} an initial state of a dishwasher  to make the behavior robust under the chaotic randomness typically exhibited in real-world (Figure \ref{dishwasher})? Where would you even start shaping things from? How would an \textit{unshaped} counterpart look like for initial/goal state shaping? 

A reasonable \textit{unshaped} intial/goal state to start with are the nominal states defined in  \textit{reference} environment -- manually designed sample environment with every actors (robots and assets) staying in its nominal pose. We often assume such nominal pose to be a mean of its underlying distribution, commonly assumed as Gaussian or Uniform distribution. This is indeed the most simple yet common technique of designing initial state distribution for many robotics tasks -- we just randomly perturb robot joints and assets around its nominal (reference) pose! 

When the task gets more complex, however, this simple approach starts to break quickly. Imagine randomly perturbing a single nominal state of a dishwasher, or randomly perturbing the nominal pose of a quadruped standing on a rough terrain; dishes and ladles, feets and terrain will be in penetration most of the time. Doing rejection sampling can be a stopgap solution, but it might end up rejecting most of the samples, making the approach nearly unusable. 

In practice, robotics engineers thus take a clever, but heavily heurstic, task-dependent approach to shape initial states. To randomly initialize a quadruped on a rough terrain, for instance, people make the robot walk off a small region of flat ground ~\cite{rudin2022learning}, letting the simulation engine figure out the phsyics constraints. To generate random initial states for cluttered bin-picking tasks, we often drop objects from height in random order. To obtain initial states to train a fall-recovery policy for humanoid, we drop the humanoid from height \cite{peng2021amp}. For in-hand manipulation tasks, to obtain a diverse set of downward-facing initial grasps to start with, we first train a grasping policy and execute it to generate diverse initial starting configurations that does not drop the object immediately \cite{chen2022system}. The task-specific nature of such strategies poses challenges in automating this shaping operation. 

Moreover, designing \textit{how} we sample from the shaped initial/goal distribution can naturally set up a learning curriculum that progresses from simpler to more challenging tasks. Take the example of training a quadruped to run at high-speed. Training only with high-speed commands could hinder the learning process. Instead, shaping a goal distribution to be a wider distribution then the actual target speed, and designing the sampling strategy to begin with slower ones and gradually increasing, can facilitate more effective learning of fast running behaviors~\cite{margolis2022rapid}.

\textbf{Terminal Condition.} ~A strict definition of \textit{unshaped} terminal condition might be to never terminate, resembling how the real world never terminates and never gets a chance to reset from the beginning. However, since this poses quite challenging problem for most learning algorithms,
one more common choice of \textit{unshaped} terminal condition that is fairly environment agnostic is to set a predefined episode time limit. 

\textit{Shaping} the terminal condition thus corresponds to deploying strategies of \textit{early termination}, which is known to significantly improve training performance~\cite{co2020ecological}. In {IsaacGymEnvs}, \texttt{Anymal} terminates when non-foot bodies contact the ground, \texttt{Humanoid} terminates when the torso falls below a certain height, and
\texttt{AllegroHand} terminates when the object falls off from the hand.
Table \ref{table:shaping-matters} shows that early termination significantly improves the training performance in all three tasks.   

\subsection{Shaping the Entire Environment is Harder than Shaping One Component} 
\label{sec:localoptima}

In the previous section, we saw that policy learning can be highly sensitive to each individual aspect of environment shaping (Table \ref{table:shaping-matters}) and that the shaping choices in IsaacGymEnvs vary qualitatively depending on the task. This motivates that, to promote the development of truly automatic behavior generation, we should consider benchmarking against a suite of unshaped environments. If an algorithm can learn policies in unshaped environments, it should be applicable to newly defined tasks and environments without requiring additional manual shaping on those environments. The unshaped environments are an appropriate benchmark for methods that couple automatic environment shaping with RL, or to directly attempt to solve by improving RL algorithms. 

\begin{figure}[t!]
\centering

\includegraphics[width=0.95\linewidth]{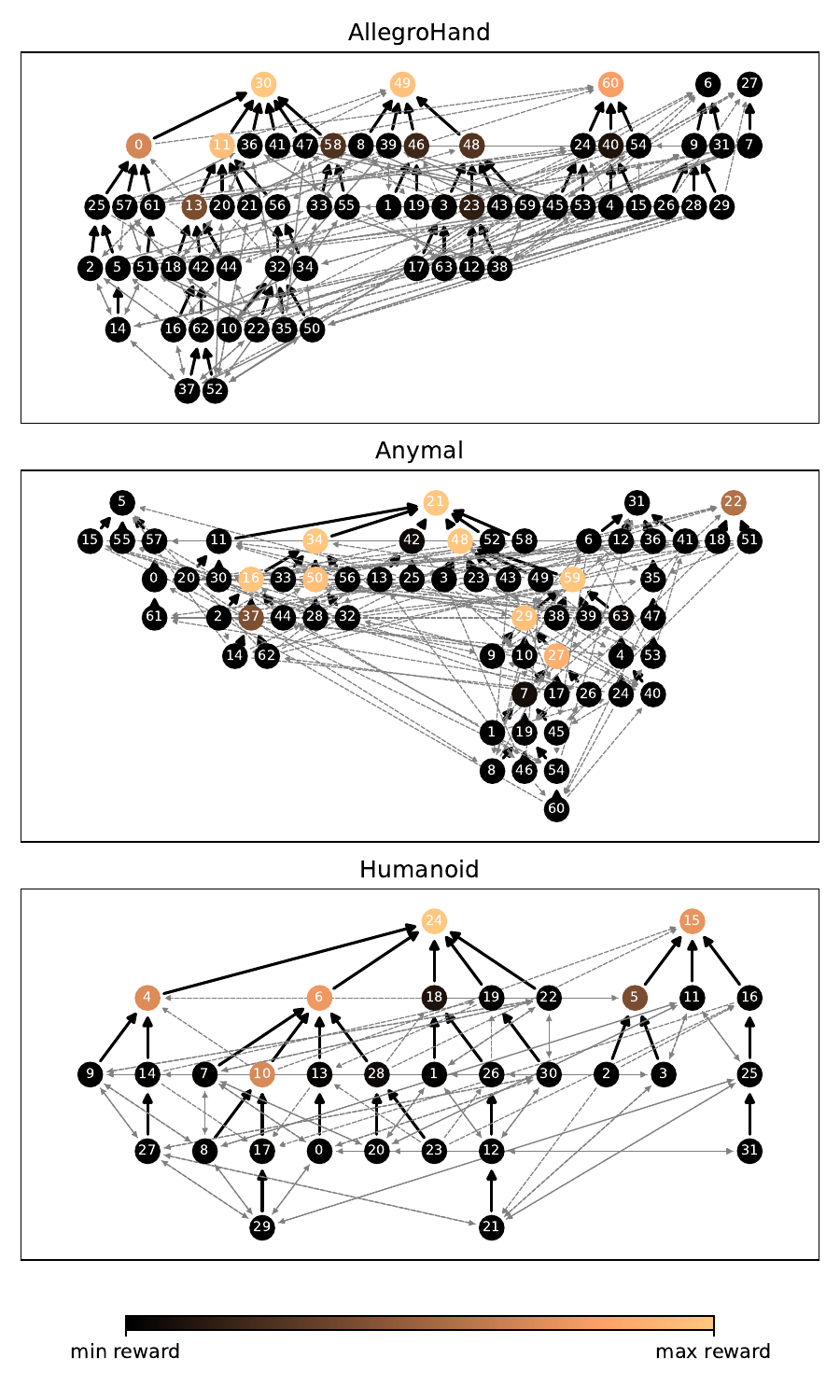}
\vspace{-0.5cm}

\caption{\textbf{Local optima in environment shaping problems.} 
Each node represents a shaped training environment. Edges connect environments that are separated by modifying one type of shaping (action space, state space, reward function, initial state, goal, or terminal condition). Bold arrows represent optimal choices for hill climbing.  Each environment is shown to have multiple local optima corresponding to the top row of nodes. 
}
\label{fig:local_optima}
\end{figure}

What kinds of methods might we try using to optimize an entirely unshaped environment? If we aim to accomplish this by introducing shaping, we may have to search over not just one aspect but all aspects of shaping. To understand the optimization landscape, suppose we have access to an oracle that proposes the human-designed environment shaping as a candidate, and we use this to perform a hill-climbing search in one aspect of the environment at a time: first, optimize the observation space, then the action space, then the reward function, and repeat until convergence.
Figure \ref{fig:local_optima} illustrates the consequence of this strategy in three environments from {IsaacGymEnvs}. Each node represents a shaped version of the environment. The node's color indicates the trained policy's performance on the shared test environment (unshaped). Edges indicate environments that differ by one type of change (state space, reward, etc.), and the bold arrows indicate the path taken by decoupled hill-climbing on one change at a time. We found that each environment has multiple local maxima where the hill climbing would get stuck in a suboptimal configuration. These local maxima differ in at least two (as many as four!) types of shaping. This suggests that the environment shaping design space is non-convex in the different types of shaping. Therefore, we should aim to develop techniques that consider shaping all parts of the environment jointly.

\subsection{Existing Automation Focuses Narrowly on Rewards}
\label{sec:eval-eureka}

There have been prior attempts to partially automate the process of environment shaping.
However, most of the efforts have focused on the subtask of reward shaping. 
The concept of reward shaping was formalized by~\cite{ng1999policy} and the idea of automated reward shaping through evolutionary search was advanced by~\citet{singh2009rewards, singh2010intrinsically}.  Recent works have formulated the problem of reward shaping as bilevel optimization with the environment design at the top level and policy learning at the bottom level and used LLMs~\cite{goyal2019using, ma2023eureka, xie2023text2reward} or gradient-based methods~\cite{hu2020learning} to generate candidates. 

LLM-based methods can benefit from prior knowledge about coding and robotics derived from internet training data to act as an efficient sampler for generating candidate shaped rewards expressed as code. Since other aspects of the environment shaping can also be expressed as code, it is straightforward to test how these methods extend. We evaluated the LLM-based reward shaping algorithm Eureka~\cite{ma2023eureka} at the task of designing the observation and action space for the \texttt{Anymal} IsaacGymEnvs environment. As in the original application to reward shaping, we generated five generations of 16 parallel candidate shaping functions using GPT-4, including the best performing candidate from each generation in the prompt for the next generation. Eureka succeeded in designing effective observation and action spaces for this environment (Table \ref{table:eureka-eval}) while all other aspects of the environment were pre-shaped. 

Motivated by Section \ref{sec:localoptima}, we also tested whether Eureka can jointly optimize multiple aspects of the environment. We found that performance drastically drops when Eureka is tasked to jointly optimize multiple environment aspects, i.e., shaping both the reward and observation, even though it could optimize each individually. This suggests that shaping multiple environment aspects jointly is yet an open problem.

\begin{table*}[]
\footnotesize
\centering
\begin{tabular}{@{}rc|c|c@{}}
\toprule
shaping component            & \begin{tabular}[c]{@{}c@{}}Eureka\\ \cite{ma2023eureka}\end{tabular} & \begin{tabular}[c]{@{}c@{}}Human Design\\ \cite{makoviychuk2021isaac}\end{tabular} & \begin{tabular}[c]{@{}l@{}}Automation\\ Performance\end{tabular} \\ \midrule
$^\star$reward              & $0.986$ & $0.973$ &  $\uparrow 0.013$  \\
$^\dag$observation          & $0.967$ & $0.973$ &  $\downarrow 0.006$          \\
$^\dag$action          & $0.982$ &   $0.973$ &   $\uparrow 0.009$                                                            \\
$^\circ$reward $\times$ observation & $0.196$  &  $0.973$ &   $\downarrow 0.777$   \\ 
$^\circ$reward $\times$ action & $0.536$  &  $0.973$ &   $\downarrow 0.437$   \\ 

$^\circ$reward $\times$ observation $\times$ action & N/A  & $0.973$ &   N/A                           \\ 
\bottomrule
\end{tabular}
\caption{Evaluating \cite{ma2023eureka} for $^\star$reward shaping, $^\dag$shaping different components, and $^\circ$coupled shaping. {\small \texttt{Anymal}} task \cite{makoviychuk2021isaac} is used as an environment. Automation performance is measured by the relative gain obtained by automating the shaping process, compared to design choices made by humans \cite{makoviychuk2021isaac}. Behavior is evaluated with the following task specification: $r = \exp\left ( {\text{negative\_distance\_to\_command}} \right )$. 5 iterations of outer loop for Eureka. Used latest GPT-4 model. Note that GPT-4 failed to generate working code for results labeled N/A. }
\label{table:eureka-eval}
\end{table*}

\label{sec:existingtools}

\section{Paths Forward to Automated Environment Shaping}

Today, no algorithm can solve diverse unshaped tasks, although we know them to be solvable with environment shaping by human designers. How can we fix this and generalize successes in RL for robotics? There are a few possibilities:

    \textbf{Scale up computation.} ~Existing bi-level search techniques like \citet{ma2023eureka} can be extended to design environment shaping and run with more compute resources to search over a greater number of candidate shaping designs. Massively parallel simulation has led to realistic robotics environments that can train quickly~\cite{makoviychuk2021isaac, rudin2022learning}; However, some state-of-the-art sim-to-real methods take weeks to train a single policy due to high data requirements or expensive subroutines~\cite{chen2023visual, jenelten2024dtc}. This would make performing much more outer-loop search impractical.

    \textbf{Improve priors.} ~If we can't search over more candidates, a better way is to generate higher-quality candidates more quickly. One possibility is that improved foundation models will zero-shot generate better candidate shaped environments~\cite{xian2023towards, yang2023holodeck}. However, it's hard to predict how much this will help. Another good idea is to \textit{mechanistically understand the strategies of human designers}. By what mechanism does the choice of observation space or curriculum improve performance? \citet{co2020ecological} proposed a holistic view of environment shaping (`Ecological Reinforcement Learning') and studied the impact of stochasticity, goal distribution, and early termination design on learnability. \citet{yu2023identifying} examined the impact of observation space on learning quadruped locomotion and offered an explanation of how different observations can work better for different subtasks. \citet{esser2023guided} surveyed a number of works across applied reinforcement learning, comparing practitioners' design choices. \citet{peng2017learning} analyzed a set of common action spaces for robot control and \citet{aljalbout2023role} developed metrics to explain the performance gap. \citet{kim2023not} found that constraints require less tuning than rewards to transfer across diverse robots. If more understanding is documented in these areas, improved biases could be implemented as a prior, e.g., in an LLM's context.

    \textbf{Shape online.} ~Instead of iteratively improving on the environment shaping $f$ across training runs, can we improve it dynamically within a RL training loop? If, for example, multi-objective reinforcement learning or bilevel optimization can trade off the coefficients of multiple reward terms~\cite{singh2010intrinsically, zheng2018learning, chen2022redeeming}, then perhaps the LLM search can be performed only over the form of the different reward terms, and their relative weighting can be optimized at runtime. In other aspects of the environment, an analogous approach would be to generate a parameterized shaping operator and automatically tune its parameters. For example, the scale of the observation or the termination height could be assigned as parameters for online optimization.

\textbf{A Robotics Benchmark for Environment Shaping.} 
Instead of evaluating RL algorithms on environments pre-shaped to work with PPO, the community should evaluate them on unshaped environments.
These environments will be \textit{too hard} to solve with existing RL algorithms. Thus, it will be necessary to (a) study how to modify aspects of the environment to make it efficiently solvable with RL and / or (b) develop better RL algorithms that can handle such challenging sparse environments. 

We modified the environments in IsaacGymEnvs to use unshaped design choices to provide a good proxy for the task-agnostic output from procedural environment generators. To facilitate environment shaping research, our code exposes an API for modifying the environment code, which allows the optimizer to transform the reward, observation space, action space, etc. by editing Python functions at runtime. The API is designed so that any language models can be easily integrated to perform such transformations. Our implementation also facilitates faster evaluation of multiple environment shaping choices by training multiple policies in a single process, leveraging parallel simulation.

\section{Conclusion}

Reinforcement learning has long promised to solve decision-making problems in a task-agnostic manner. It has found great success in solving challenging but narrowly-scoped tasks in robotics. In this position paper, we argued that the key bottleneck for scalability of RL is a limited mechanistic understanding of task-specific engineering (\textit{environment shaping}) that transforms environments to be solved more easily and is universal across domains and benchmarks. We proposed a formal definition of environment shaping as an optimization problem and identified instances of shaping in robotic tasks. Finally, we identified key steps forward such as developing computationally efficient search over environment shaping; improving our tools for implementing such shaping and understanding its impact on learning dynamics; and defining benchmarks for this problem. We hope this will motivate an increased focus in RL research on communicating and evaluating environment-shaping measures that impact performance rather than solely emphasizing the impact of the learning algorithm or policy architecture.

\section*{Acknowledgements}

We thank the members of the Improbable AI lab for the helpful discussions and feedback on the paper. We are grateful to MIT Supercloud and the Lincoln Laboratory Supercomputing Center for providing HPC resources. This research was partly supported by Hyundai Motor Company, DARPA Machine Common Sense Program, the MIT-IBM Watson AI Lab, and the National Science Foundation under Cooperative Agreement PHY-2019786 (The NSF AI Institute for Artificial Intelligence and Fundamental Interactions, http://iaifi.org/). We acknowledge support from ONR MURI under grant number N00014-22-1-2740. Research was sponsored by the Army Research Office and was accomplished under Grant Number W911NF-21-1-0328. The views and conclusions contained in this document are those of the authors and should not be interpreted as representing the official policies, either expressed or implied, of the Army Research Office or the U.S. Government. The U.S. Government is authorized to reproduce and distribute reprints for Government purposes notwithstanding any copyright notation herein.

\section*{Impact Statement}

This paper presents work whose goal is to advance the field of 
Machine Learning. There are many potential societal consequences 
of our work, none which we feel must be specifically highlighted here.

\bibliography{example_paper}

\begin{thebibliography}{59}
\providecommand{\natexlab}[1]{#1}
\providecommand{\url}[1]{\texttt{#1}}
\expandafter\ifx\csname urlstyle\endcsname\relax
  \providecommand{\doi}[1]{doi: #1}\else
  \providecommand{\doi}{doi: \begingroup \urlstyle{rm}\Url}\fi

\bibitem[Agrawal(2022)]{task-specification-problem}
Agrawal, P.
\newblock The task specification problem.
\newblock In \emph{Conference on Robot Learning}, pp.\  1745--1751. PMLR, 2022.

\bibitem[Aljalbout et~al.(2023)Aljalbout, Frank, Karl, and van~der Smagt]{aljalbout2023role}
Aljalbout, E., Frank, F., Karl, M., and van~der Smagt, P.
\newblock On the role of the action space in robot manipulation learning and sim-to-real transfer.
\newblock \emph{arXiv preprint arXiv:2312.03673}, 2023.

\bibitem[Andrychowicz et~al.(2020)Andrychowicz, Baker, Chociej, Jozefowicz, McGrew, Pachocki, Petron, Plappert, Powell, Ray, et~al.]{andrychowicz2020learning}
Andrychowicz, O.~M., Baker, B., Chociej, M., Jozefowicz, R., McGrew, B., Pachocki, J., Petron, A., Plappert, M., Powell, G., Ray, A., et~al.
\newblock Learning dexterous in-hand manipulation.
\newblock \emph{The International Journal of Robotics Research}, 39\penalty0 (1):\penalty0 3--20, 2020.

\bibitem[Chen et~al.(2020)Chen, Zhou, Koltun, and Kr{\"a}henb{\"u}hl]{chen2020learning}
Chen, D., Zhou, B., Koltun, V., and Kr{\"a}henb{\"u}hl, P.
\newblock Learning by cheating.
\newblock In \emph{Conference on Robot Learning}, pp.\  66--75. PMLR, 2020.

\bibitem[Chen et~al.(2022{\natexlab{a}})Chen, Hong, Pajarinen, and Agrawal]{chen2022redeeming}
Chen, E., Hong, Z.-W., Pajarinen, J., and Agrawal, P.
\newblock Redeeming intrinsic rewards via constrained optimization.
\newblock \emph{Advances in Neural Information Processing Systems}, 35:\penalty0 4996--5008, 2022{\natexlab{a}}.

\bibitem[Chen et~al.(2022{\natexlab{b}})Chen, Xu, and Agrawal]{chen2022system}
Chen, T., Xu, J., and Agrawal, P.
\newblock A system for general in-hand object re-orientation.
\newblock In \emph{Conference on Robot Learning}, pp.\  297--307. PMLR, 2022{\natexlab{b}}.

\bibitem[Chen et~al.(2023)Chen, Tippur, Wu, Kumar, Adelson, and Agrawal]{chen2023visual}
Chen, T., Tippur, M., Wu, S., Kumar, V., Adelson, E., and Agrawal, P.
\newblock Visual dexterity: In-hand reorientation of novel and complex object shapes.
\newblock \emph{Science Robotics}, 8\penalty0 (84):\penalty0 eadc9244, 2023.

\bibitem[Cheng et~al.(2023)Cheng, Shi, Agarwal, and Pathak]{cheng2023extreme}
Cheng, X., Shi, K., Agarwal, A., and Pathak, D.
\newblock Extreme parkour with legged robots.
\newblock \emph{arXiv preprint arXiv:2309.14341}, 2023.

\bibitem[Co-Reyes et~al.(2020)Co-Reyes, Sanjeev, Berseth, Gupta, and Levine]{co2020ecological}
Co-Reyes, J.~D., Sanjeev, S., Berseth, G., Gupta, A., and Levine, S.
\newblock Ecological reinforcement learning.
\newblock \emph{arXiv preprint arXiv:2006.12478}, 2020.

\bibitem[Eschmann(2021)]{eschmann2021reward}
Eschmann, J.
\newblock Reward function design in reinforcement learning.
\newblock \emph{Reinforcement Learning Algorithms: Analysis and Applications}, pp.\  25--33, 2021.

\bibitem[E{\ss}er et~al.(2023)E{\ss}er, Bach, Jestel, Urbann, and Kerner]{esser2023guided}
E{\ss}er, J., Bach, N., Jestel, C., Urbann, O., and Kerner, S.
\newblock Guided reinforcement learning: A review and evaluation for efficient and effective real-world robotics [survey].
\newblock \emph{IEEE Robotics \& Automation Magazine}, 30\penalty0 (2):\penalty0 67--85, 2023.
\newblock \doi{10.1109/MRA.2022.3207664}.

\bibitem[Fu et~al.(2024)Fu, Zhao, and Finn]{MobileALOHA}
Fu, Z., Zhao, T.~Z., and Finn, C.
\newblock Mobile aloha: Learning bimanual mobile manipulation with low-cost whole-body teleoperation.
\newblock \emph{arXiv preprint arXiv:2401.02117}, 2024.

\bibitem[Goyal et~al.(2019)Goyal, Niekum, and Mooney]{goyal2019using}
Goyal, P., Niekum, S., and Mooney, R.~J.
\newblock Using natural language for reward shaping in reinforcement learning.
\newblock \emph{arXiv preprint arXiv:1903.02020}, 2019.

\bibitem[Gupta et~al.(2022)Gupta, Pacchiano, Zhai, Kakade, and Levine]{gupta2022unpacking}
Gupta, A., Pacchiano, A., Zhai, Y., Kakade, S.~M., and Levine, S.
\newblock Unpacking reward shaping: Understanding the benefits of reward engineering on sample complexity, 2022.

\bibitem[Handa et~al.(2023)Handa, Allshire, Makoviychuk, Petrenko, Singh, Liu, Makoviichuk, Van~Wyk, Zhurkevich, Sundaralingam, et~al.]{handa2023dextreme}
Handa, A., Allshire, A., Makoviychuk, V., Petrenko, A., Singh, R., Liu, J., Makoviichuk, D., Van~Wyk, K., Zhurkevich, A., Sundaralingam, B., et~al.
\newblock Dextreme: Transfer of agile in-hand manipulation from simulation to reality.
\newblock In \emph{2023 IEEE International Conference on Robotics and Automation (ICRA)}, pp.\  5977--5984. IEEE, 2023.

\bibitem[Hoeller et~al.(2023)Hoeller, Rudin, Sako, and Hutter]{hoeller2023anymal}
Hoeller, D., Rudin, N., Sako, D., and Hutter, M.
\newblock Anymal parkour: Learning agile navigation for quadrupedal robots.
\newblock \emph{arXiv preprint arXiv:2306.14874}, 2023.

\bibitem[Hogan(1984)]{hogan1984impedance}
Hogan, N.
\newblock Impedance control: An approach to manipulation.
\newblock In \emph{1984 American control conference}, pp.\  304--313. IEEE, 1984.

\bibitem[Hu et~al.(2020)Hu, Wang, Jia, Wang, Chen, Hao, Wu, and Fan]{hu2020learning}
Hu, Y., Wang, W., Jia, H., Wang, Y., Chen, Y., Hao, J., Wu, F., and Fan, C.
\newblock Learning to utilize shaping rewards: A new approach of reward shaping.
\newblock \emph{Advances in Neural Information Processing Systems}, 33:\penalty0 15931--15941, 2020.

\bibitem[Jenelten et~al.(2024)Jenelten, He, Farshidian, and Hutter]{jenelten2024dtc}
Jenelten, F., He, J., Farshidian, F., and Hutter, M.
\newblock Dtc: Deep tracking control.
\newblock \emph{Science Robotics}, 9\penalty0 (86):\penalty0 eadh5401, 2024.

\bibitem[Ji et~al.(2023)Ji, Margolis, and Agrawal]{ji2023dribble}
Ji, Y., Margolis, G.~B., and Agrawal, P.
\newblock Dribblebot: Dynamic legged manipulation in the wild.
\newblock \emph{International Conference on Robotics and Automation}, 2023.

\bibitem[Kaufmann et~al.(2023)Kaufmann, Bauersfeld, Loquercio, M{\"u}ller, Koltun, and Scaramuzza]{kaufmann2023champion}
Kaufmann, E., Bauersfeld, L., Loquercio, A., M{\"u}ller, M., Koltun, V., and Scaramuzza, D.
\newblock Champion-level drone racing using deep reinforcement learning.
\newblock \emph{Nature}, 620\penalty0 (7976):\penalty0 982--987, 2023.

\bibitem[Khatib(1985)]{khatib1985operational}
Khatib, O.
\newblock The operational space formulation in the analysis, design, and control of robot manipulators.
\newblock In \emph{Preprints 3rd International Symposium of Robotics Re-search, Gouvieux (Chantilly), France, October}, pp.\  7--11, 1985.

\bibitem[Kim et~al.(2023)Kim, Oh, Lee, Choi, Ji, Jung, Youm, and Hwangbo]{kim2023not}
Kim, Y., Oh, H., Lee, J., Choi, J., Ji, G., Jung, M., Youm, D., and Hwangbo, J.
\newblock Not only rewards but also constraints: Applications on legged robot locomotion.
\newblock \emph{arXiv preprint arXiv:2308.12517}, 2023.

\bibitem[Kiran \& Ozyildirim(2022)Kiran and Ozyildirim]{kiran2022hyperparameter}
Kiran, M. and Ozyildirim, M.
\newblock Hyperparameter tuning for deep reinforcement learning applications.
\newblock \emph{arXiv preprint arXiv:2201.11182}, 2022.

\bibitem[Lee et~al.(2020)Lee, Hwangbo, Wellhausen, Koltun, and Hutter]{lee2020learning}
Lee, J., Hwangbo, J., Wellhausen, L., Koltun, V., and Hutter, M.
\newblock Learning quadrupedal locomotion over challenging terrain.
\newblock \emph{Science robotics}, 5\penalty0 (47):\penalty0 eabc5986, 2020.

\bibitem[Lei et~al.(2023)Lei, He, Lu, Hu, Gao, and Xu]{lei2023uni}
Lei, K., He, Z., Lu, C., Hu, K., Gao, Y., and Xu, H.
\newblock Uni-o4: Unifying online and offline deep reinforcement learning with multi-step on-policy optimization.
\newblock \emph{arXiv preprint arXiv:2311.03351}, 2023.

\bibitem[Liu et~al.(2021)Liu, Rui, Gao, Yang, Yang, Wang, Wang, and Jian]{finrl_meta_2021}
Liu, X.-Y., Rui, J., Gao, J., Yang, L., Yang, H., Wang, Z., Wang, C.~D., and Jian, G.
\newblock {FinRL-Meta}: Data-driven deep reinforcementlearning in quantitative finance.
\newblock \emph{Data-Centric AI Workshop, NeurIPS}, 2021.

\bibitem[Luo et~al.(2024)Luo, Hu, Xu, Tan, Berg, Sharma, Schaal, Finn, Gupta, and Levine]{luo2024serl}
Luo, J., Hu, Z., Xu, C., Tan, Y.~L., Berg, J., Sharma, A., Schaal, S., Finn, C., Gupta, A., and Levine, S.
\newblock Serl: A software suite for sample-efficient robotic reinforcement learning.
\newblock \emph{arXiv preprint arXiv:2401.16013}, 2024.

\bibitem[Ma et~al.(2023)Ma, Liang, Wang, Huang, Bastani, Jayaraman, Zhu, Fan, and Anandkumar]{ma2023eureka}
Ma, Y.~J., Liang, W., Wang, G., Huang, D.-A., Bastani, O., Jayaraman, D., Zhu, Y., Fan, L., and Anandkumar, A.
\newblock Eureka: Human-level reward design via coding large language models.
\newblock 2023.

\bibitem[Makoviychuk et~al.(2021)Makoviychuk, Wawrzyniak, Guo, Lu, Storey, Macklin, Hoeller, Rudin, Allshire, Handa, and State]{makoviychuk2021isaac}
Makoviychuk, V., Wawrzyniak, L., Guo, Y., Lu, M., Storey, K., Macklin, M., Hoeller, D., Rudin, N., Allshire, A., Handa, A., and State, G.
\newblock Isaac gym: High performance gpu-based physics simulation for robot learning, 2021.

\bibitem[Margolis \& Agrawal(2023)Margolis and Agrawal]{margolis2023walk}
Margolis, G.~B. and Agrawal, P.
\newblock Walk these ways: Tuning robot control for generalization with multiplicity of behavior.
\newblock In \emph{Conference on Robot Learning}, pp.\  22--31. PMLR, 2023.

\bibitem[Margolis et~al.(2022)Margolis, Yang, Paigwar, Chen, and Agrawal]{margolis2022rapid}
Margolis, G.~B., Yang, G., Paigwar, K., Chen, T., and Agrawal, P.
\newblock Rapid locomotion via reinforcement learning.
\newblock \emph{arXiv preprint arXiv:2205.02824}, 2022.

\bibitem[Matiisen et~al.(2019)Matiisen, Oliver, Cohen, and Schulman]{matiisen2019teacher}
Matiisen, T., Oliver, A., Cohen, T., and Schulman, J.
\newblock Teacher--student curriculum learning.
\newblock \emph{IEEE transactions on neural networks and learning systems}, 31\penalty0 (9):\penalty0 3732--3740, 2019.

\bibitem[Miki et~al.(2022)Miki, Lee, Hwangbo, Wellhausen, Koltun, and Hutter]{miki2022learning}
Miki, T., Lee, J., Hwangbo, J., Wellhausen, L., Koltun, V., and Hutter, M.
\newblock Learning robust perceptive locomotion for quadrupedal robots in the wild.
\newblock \emph{Science Robotics}, 7\penalty0 (62):\penalty0 eabk2822, 2022.

\bibitem[Ng et~al.(1999)Ng, Harada, and Russell]{ng1999policy}
Ng, A.~Y., Harada, D., and Russell, S.
\newblock Policy invariance under reward transformations: Theory and application to reward shaping.
\newblock In \emph{Icml}, volume~99, pp.\  278--287. Citeseer, 1999.

\bibitem[Niekum et~al.(2010)Niekum, Barto, and Spector]{niekum2010genetic}
Niekum, S., Barto, A.~G., and Spector, L.
\newblock Genetic programming for reward function search.
\newblock \emph{IEEE Transactions on Autonomous Mental Development}, 2\penalty0 (2):\penalty0 83--90, 2010.

\bibitem[Parker-Holder et~al.(2022)Parker-Holder, Rajan, Song, Biedenkapp, Miao, Eimer, Zhang, Nguyen, Calandra, Faust, et~al.]{parker2022automated}
Parker-Holder, J., Rajan, R., Song, X., Biedenkapp, A., Miao, Y., Eimer, T., Zhang, B., Nguyen, V., Calandra, R., Faust, A., et~al.
\newblock Automated reinforcement learning (autorl): A survey and open problems.
\newblock \emph{Journal of Artificial Intelligence Research}, 74:\penalty0 517--568, 2022.

\bibitem[Pathak et~al.(2017)Pathak, Agrawal, Efros, and Darrell]{pathak2017curiosity}
Pathak, D., Agrawal, P., Efros, A.~A., and Darrell, T.
\newblock Curiosity-driven exploration by self-supervised prediction.
\newblock In \emph{International conference on machine learning}, pp.\  2778--2787. PMLR, 2017.

\bibitem[Peng \& Van De~Panne(2017)Peng and Van De~Panne]{peng2017learning}
Peng, X.~B. and Van De~Panne, M.
\newblock Learning locomotion skills using deeprl: Does the choice of action space matter?
\newblock In \emph{Proceedings of the ACM SIGGRAPH/Eurographics Symposium on Computer Animation}, pp.\  1--13, 2017.

\bibitem[Peng et~al.(2021)Peng, Ma, Abbeel, Levine, and Kanazawa]{peng2021amp}
Peng, X.~B., Ma, Z., Abbeel, P., Levine, S., and Kanazawa, A.
\newblock Amp: Adversarial motion priors for stylized physics-based character control.
\newblock \emph{ACM Transactions on Graphics (ToG)}, 40\penalty0 (4):\penalty0 1--20, 2021.

\bibitem[Portelas et~al.(2020)Portelas, Colas, Weng, Hofmann, and Oudeyer]{portelas2020automatic}
Portelas, R., Colas, C., Weng, L., Hofmann, K., and Oudeyer, P.-Y.
\newblock Automatic curriculum learning for deep rl: A short survey.
\newblock \emph{arXiv preprint arXiv:2003.04664}, 2020.

\bibitem[Rudin et~al.(2022)Rudin, Hoeller, Reist, and Hutter]{rudin2022learning}
Rudin, N., Hoeller, D., Reist, P., and Hutter, M.
\newblock Learning to walk in minutes using massively parallel deep reinforcement learning.
\newblock In \emph{Conference on Robot Learning}, pp.\  91--100. PMLR, 2022.

\bibitem[Schulman et~al.(2017)Schulman, Wolski, Dhariwal, Radford, and Klimov]{schulman2017proximal}
Schulman, J., Wolski, F., Dhariwal, P., Radford, A., and Klimov, O.
\newblock Proximal policy optimization algorithms.
\newblock \emph{arXiv preprint arXiv:1707.06347}, 2017.

\bibitem[Selfridge et~al.(1985)Selfridge, Sutton, and Barto]{selfridge1985training}
Selfridge, O.~G., Sutton, R.~S., and Barto, A.~G.
\newblock Training and tracking in robotics.
\newblock In \emph{Ijcai}, pp.\  670--672, 1985.

\bibitem[Singh et~al.(2009)Singh, Lewis, and Barto]{singh2009rewards}
Singh, S., Lewis, R.~L., and Barto, A.~G.
\newblock Where do rewards come from.
\newblock In \emph{Proceedings of the annual conference of the cognitive science society}, pp.\  2601--2606. Cognitive Science Society, 2009.

\bibitem[Singh et~al.(2010)Singh, Lewis, Barto, and Sorg]{singh2010intrinsically}
Singh, S., Lewis, R.~L., Barto, A.~G., and Sorg, J.
\newblock Intrinsically motivated reinforcement learning: An evolutionary perspective.
\newblock \emph{IEEE Transactions on Autonomous Mental Development}, 2\penalty0 (2):\penalty0 70--82, 2010.

\bibitem[Song et~al.(2023)Song, Romero, M{\"u}ller, Koltun, and Scaramuzza]{song2023reaching}
Song, Y., Romero, A., M{\"u}ller, M., Koltun, V., and Scaramuzza, D.
\newblock Reaching the limit in autonomous racing: Optimal control versus reinforcement learning.
\newblock \emph{Science Robotics}, 8\penalty0 (82):\penalty0 eadg1462, 2023.

\bibitem[V{\'a}zquez-Canteli et~al.(2019)V{\'a}zquez-Canteli, K{\"a}mpf, Henze, and Nagy]{vazquez2019citylearn}
V{\'a}zquez-Canteli, J.~R., K{\"a}mpf, J., Henze, G., and Nagy, Z.
\newblock Citylearn v1. 0: An openai gym environment for demand response with deep reinforcement learning.
\newblock In \emph{Proceedings of the 6th ACM international conference on systems for energy-efficient buildings, cities, and transportation}, pp.\  356--357, 2019.

\bibitem[Wei et~al.(2018)Wei, Zheng, Yao, and Li]{wei2018intellilight}
Wei, H., Zheng, G., Yao, H., and Li, Z.
\newblock Intellilight: A reinforcement learning approach for intelligent traffic light control.
\newblock In \emph{Proceedings of the 24th ACM SIGKDD international conference on knowledge discovery \& data mining}, pp.\  2496--2505, 2018.

\bibitem[Wu et~al.(2017)Wu, Kreidieh, Vinitsky, and Bayen]{wu2017emergent}
Wu, C., Kreidieh, A., Vinitsky, E., and Bayen, A.~M.
\newblock Emergent behaviors in mixed-autonomy traffic.
\newblock In \emph{Conference on Robot Learning}, pp.\  398--407. PMLR, 2017.

\bibitem[Wu et~al.(2023)Wu, Escontrela, Hafner, Abbeel, and Goldberg]{wu2023daydreamer}
Wu, P., Escontrela, A., Hafner, D., Abbeel, P., and Goldberg, K.
\newblock Daydreamer: World models for physical robot learning.
\newblock In \emph{Conference on Robot Learning}, pp.\  2226--2240. PMLR, 2023.

\bibitem[Xian et~al.(2023)Xian, Gervet, Xu, Qiao, and Wang]{xian2023towards}
Xian, Z., Gervet, T., Xu, Z., Qiao, Y.-L., and Wang, T.-H.
\newblock Towards a foundation model for generalist robots: Diverse skill learning at scale via automated task and scene generation.
\newblock \emph{arXiv preprint arXiv:2305.10455}, 2023.

\bibitem[Xie et~al.(2023)Xie, Zhao, Wu, Liu, Luo, Zhong, Yang, and Yu]{xie2023text2reward}
Xie, T., Zhao, S., Wu, C.~H., Liu, Y., Luo, Q., Zhong, V., Yang, Y., and Yu, T.
\newblock Text2reward: Automated dense reward function generation for reinforcement learning.
\newblock \emph{arXiv preprint arXiv:2309.11489}, 2023.

\bibitem[Yang et~al.(2023)Yang, Sun, Weihs, VanderBilt, Herrasti, Han, Wu, Haber, Krishna, Liu, et~al.]{yang2023holodeck}
Yang, Y., Sun, F.-Y., Weihs, L., VanderBilt, E., Herrasti, A., Han, W., Wu, J., Haber, N., Krishna, R., Liu, L., et~al.
\newblock Holodeck: Language guided generation of 3d embodied ai environments.
\newblock \emph{arXiv preprint arXiv:2312.09067}, 2023.

\bibitem[Yin et~al.(2023)Yin, Huang, Qin, Chen, and Wang]{yin2023rotating}
Yin, Z.-H., Huang, B., Qin, Y., Chen, Q., and Wang, X.
\newblock Rotating without seeing: Towards in-hand dexterity through touch.
\newblock \emph{arXiv preprint arXiv:2303.10880}, 2023.

\bibitem[Yu et~al.(2023)Yu, Yang, McGreavy, Triantafyllidis, Bellegarda, Shafiee, Ijspeert, and Li]{yu2023identifying}
Yu, W., Yang, C., McGreavy, C., Triantafyllidis, E., Bellegarda, G., Shafiee, M., Ijspeert, A.~J., and Li, Z.
\newblock Identifying important sensory feedback for learning locomotion skills.
\newblock \emph{Nature Machine Intelligence}, 5\penalty0 (8):\penalty0 919--932, 2023.

\bibitem[Zhao et~al.(2023)Zhao, Kumar, Levine, and Finn]{ALOHA}
Zhao, T.~Z., Kumar, V., Levine, S., and Finn, C.
\newblock Learning fine-grained bimanual manipulation with low-cost hardware.
\newblock \emph{arXiv preprint arXiv:2304.13705}, 2023.

\bibitem[Zheng et~al.(2018)Zheng, Oh, and Singh]{zheng2018learning}
Zheng, Z., Oh, J., and Singh, S.
\newblock On learning intrinsic rewards for policy gradient methods.
\newblock \emph{Advances in Neural Information Processing Systems}, 31, 2018.

\bibitem[Zhuang et~al.(2023)Zhuang, Fu, Wang, Atkeson, Schwertfeger, Finn, and Zhao]{zhuang2023robot}
Zhuang, Z., Fu, Z., Wang, J., Atkeson, C., Schwertfeger, S., Finn, C., and Zhao, H.
\newblock Robot parkour learning.
\newblock \emph{arXiv preprint arXiv:2309.05665}, 2023.

\end{thebibliography}
\bibliographystyle{icml2024}

\end{document}